\documentclass[letterpaper, 10 pt, conference]{ieeeconf}  

\IEEEoverridecommandlockouts

\usepackage{graphicx}
\usepackage{caption}
\usepackage{subcaption}

\usepackage{algorithm}
\usepackage[noend]{algpseudocode}
\usepackage{amsmath}
\usepackage{comment}
\usepackage{amssymb}

\usepackage[backend=biber,style=numeric,sorting=none]{biblatex}
\title{\LARGE \bf
Imaginary Hindsight Experience Replay: Curious Model-based Learning for Sparse Reward Tasks
}

\author{Robert McCarthy$^{\dagger}$, Qiang Wang$^{\dagger}$, and Stephen J. Redmond$^{\dagger}$
\thanks{$^{\dagger}$UCD School of Electronic and Electrical Engineering, University College Dublin, Belfield, Dublin 4, Ireland. e-mail: \{robert.mccarthy, qiang.wang\}@ucdconnect.ie and stephen.redmond@ucd.ie.}
}

\begin{document}

\maketitle
\thispagestyle{empty}
\pagestyle{empty}

%%%%%%%%%%%%%%%%%%%%%%%%%%%%%%%%%%%%%%%%%%%%%%%%%%%%%%%%%%%%%%%%%%%%%%%%%%%%%%%%
\begin{abstract}

Model-based reinforcement learning is a promising learning strategy for practical robotic applications due to its improved data-efficiency versus model-free counterparts. However, current state-of-the-art model-based methods rely on shaped reward signals, which can be difficult to design and implement. To remedy this, we propose a simple model-based method tailored for sparse-reward multi-goal tasks that foregoes the need for complicated reward engineering. This approach, termed Imaginary Hindsight Experience Replay, minimises real-world interactions by incorporating imaginary data into policy updates. To improve exploration in the sparse-reward setting, the policy is trained with standard Hindsight Experience Replay and endowed with curiosity-based intrinsic rewards. Upon evaluation, this approach provides an order of magnitude increase in data-efficiency on average versus the state-of-the-art model-free method in the benchmark OpenAI Gym Fetch Robotics tasks.

\end{abstract}

%%%%%%%%%%%%%%%%%%%%%%%%%%%%%%%%%%%%%%%%%%%%%%%%%%%%%%%%%%%%%%%%%%%%%%%%%%%%%%%%
\section{INTRODUCTION}

Deep reinforcement learning (RL) has achieved remarkable successes in recent years, for example learning to play Atari computer games \cite{mnih2015human} or defeating the world champion in the game of Dota 2 \cite{berner2019dota}. However, these successes have generally relied on the availability of vast amounts of training data and a dense/well-shaped reward function to guide exploration, neither of which are easily obtained in practical robotic scenarios. Thus, while RL has impressed in virtual worlds, successes in the real world have been relatively limited.

% That RL is so data hungry is a fundamental barrier to its application in robotics; unlike in video games, gathering data in the real world is prohibitively time-consuming and expensive. To address this issue, imitation learning has aimed to reduce or completely eliminate the exploration phase required by the agent by providing it with expert trajectories to learn from. However, an expert may not always be available to generate such exemplars. Meanwhile, simulation-to-real (sim-to-real) transfer methods learn the policy in simulation before transferring it to the real world for fine-tuning and/or evaluation. This can allow the agent to learn from the real-world equivalent of years of (simulated) experience in a matter of hours \cite{andrychowicz2020learning}, but a significant engineering effort is required to create a high fidelity simulator, especially as the complexity of the environment grows; for example, simulating the contact mechanics of robotics applications is non-trivial.

The data hungry nature of RL is a fundamental barrier to its application in robotics; unlike in video games, gathering data in the real world is prohibitively time-consuming and expensive. However, model-based RL has shown promise in improving sample efficiency and tackling this issue. Unlike imitation learning \cite{hussein2017imitation} or simulation to real transfer (sim-to-real) \cite{andrychowicz2020learning}, these algorithms do not come with a hard requirement for expert domain knowledge or significant engineering effort. Unlike model-free RL methods, model-based RL learns to explicitly model the transition dynamics of the environment, using this model to aid the learning of a control policy. Since the model can be learned in an entirely supervised manner, the primary advantage of model-based RL is a significant improvement in sample efficiency. In fact, recent advances in the state-of-the-art have allowed model-based methods to match the asymptotic performance of model-free methods, while requiring orders of magnitude less data \cite{janner2019trust,hafner2020mastering,schrittwieser2020mastering}. However, data-efficiency issues aside, these state-of-the-art techniques generally rely on dense reward signals, and in this respect are still somewhat impractical for robotics applications.

In robotic manipulation tasks, significant expertise can be required to design and implement a dense reward that can successfully guide learning. For example, Popov et al. \cite{popov2017data} require a highly complex reward function to learn to stack blocks, while Andrychowicz et al. \cite{andrychowicz2017hindsight} demonstrate that an over-simplified dense reward function can be detrimental to learning performances. Suitable shaped rewards are not only difficult to design, but also difficult to implement in the real world; unlike in simulation, often computer vision systems must be employed to track reward-relevant information, such as object positions \cite{nagabandi2020deep}.  

To bypass the challenge of hand-crafting dense reward signals, methods have been developed which can learn from sparse rewards; rewards with little information that are generally only received once the task has been successfully completed. These sparse rewards are much easier to design and implement, but are conversely much more difficult to learn from. Deep RL methods such as curiosity-driven exploration \cite{pathak2017curiosity,burda2018exploration} and Hindsight Experience Replay (HER) \cite{andrychowicz2017hindsight} have made significant progress in the sparse-reward setting, however, few works have successfully integrated these techniques with model-based methods to obtain the benefits of both in difficult robotic tasks.

% To help the agent uncover such sparse rewards, it should be endowed with a useful exploration strategy. In this regard, curiosity-driven exploration methods have proven beneficial \cite{pathak2017curiosity,burda2018exploration}, by encouraging the agent to visit novel states and efficiently cover the state-space. Meanwhile, Hindsight Experience Replay (HER) \cite{andrychowicz2017hindsight} accelerates exploration by allowing the agent to learn from what was achieved in its failed training episodes. HER is particularly well suited to the object manipulation tasks commonly found in robotics, and has often proved essential for learning these tasks via RL without excessive reward engineering. 

% To help the agent to obtain these sparse rewards, it should be endowed with a useful exploration strategy. Recently, in deep RL, curiosity-driven exploration methods have shown themselves to be one such strategy \cite{pathak2017curiosity,burda2018exploration}. Curiosity-drive exploration encourages the agent to visit novel system states, allowing it to efficiently explore the state space, gaining useful skills and knowledge about the environment in the process. Another technique used to learn in sparse reward tasks is Hindsight Experience Replay (HER) \cite{andrychowicz2017hindsight}. HER accelerates exploration by allowing the agent to learn from what was achieved in its failed training episodes. HER is particularly suited to object manipulation tasks commonly found in robotics, and has often proved essential for learning these tasks via RL.

In this paper, a novel model-based RL method specifically tailored for sparse-reward tasks is proposed to jointly address the data-efficiency and reward engineering issues inherent in the application of RL to robotics. The method trains the policy with HER to ensure maximum efficacy in the multi-goal tasks commonly found in robotics, and incorporates imaginary/model-generated data into policy updates to reduce sample complexity. To ensure the learned policy does not overfit to model inaccuracies, imagined data is generated by an ensemble of models \cite{kurutach2018model}, and is regenerated each time the model is updated. To improve environment exploration, the ensemble also provides the agent with intrinsic rewards based on the disagreement between its predictions \cite{pathak2019self}. To allow the agent to adapt its behaviors specifically for the real environment, real and imaginary data are distinguished from each other when input to the policy. We term this technique Imaginary Hindsight Experience Replay (I-HER).

% To avoid the adverse effects of learning from imaginary data generated by an older, more inaccurate model ensemble, this imaginary data is rapidly regenerated each time the model is updated.

I-HER is evaluated in the challenging OpenAI Gym Fetch Robotics tasks \cite{brockman2016openai,plappert2018multi}, and converges using, on average, an order of magnitude less data than the state-of-the-art model-free algorithm, matching its asymptotic performance in all but one task. The diversity of the Fetch tasks is found to expose various modes in which model-based RL can struggle, but an ablation study demonstrates how I-HER's components help to overcome these issues. Compared to state-of-the-art model-free and model-based methods, I-HER offers increased feasibility for robotic applications due to its combined data-efficiency and ability to effectively tackle sparse-reward tasks. Moreover, I-HER's empirical results outperform those presented by other model-based methods also tailored for multi-goal sparse-reward tasks. 

% I-HER is increasingly feasible for robotic applications versus previous model-free and model-based RL techniques

\section{Related Work}

\subsection{Model-based Reinforcement Learning}

The data-efficiency of model-based RL methods make them promising for real-world applications, and many variants have been proposed. Learned dynamics models have been used for model predictive control (MPC), but these methods have struggled to match the asymptotic performance of model-free methods \cite{nagabandi2018neural}, and generally rely on a dense reward function to identify good trajectories \cite{nagabandi2020deep,hafner2019learning,chua2018deep}. While MPC is purely model-based, often model-based and model-free components are combined to yield both high data efficiency and high asymptotic performance. In one such approach, a dynamics model can be used to predict into the future to improve the value estimates of a model-free algorithm \cite{feinberg2018model,buckman2018sample}. Another approach is to train a model-free policy with cheap model-generated data; a concept first introduced by Dyna-Q \cite{sutton1990integrated}.

Like any model-based method, Dyna-Q-type approaches are limited by inaccuracies in the learned dynamics model \cite{gu2016continuous, kaiser2019model}. Kalweit and Boedecker \cite{kalweit2017uncertainty} prevent an inaccurate model from limiting asymptotic performance by sampling less imaginary data as uncertainty in the Q-function decreases. Other works employ an ensemble of dynamics models to prevent the policy from overfitting to the inaccuracies of a single model \cite{kurutach2018model, clavera2018model}. SLBO \cite{luo2018algorithmic} minimises overfitting to any single iteration of the model by updating the model regularly. MBPO \cite{janner2019trust} optimises the length of model-generated rollouts to avoid compounding model errors, while Ha and Schmidhuber \cite{ha2018world} optimise stochasticity in the imagined environment to prevent exploitation of a deterministic model.

This paper extends these Dyna-Q approaches, accounting for model inaccuracies by (i) employing an ensemble of models, (ii) introducing a step which regenerates the data in the imaginary replay buffer each time the model is updated so as to eradicate inaccuracies produced by older iterations of the model (unlike in \cite{kalweit2017uncertainty}), and (iii) introducing an explicit distinguishing of real and imaginary transitions to allow the policy to account for differences in their dynamics. 

% While these Dyna-Q approaches simply use the model as a derivative-free simulator, another promising approach is to leverage the gradients offered by the neural network dynamics. This approach, which involves back-propagating through imagined trajectories, can suffer from exploding and diminishing gradients \cite{kurutach2018model} but has recently seen great success using short imagined rollouts and an actor critic algorithm which accounts for rewards beyond the imagination horizon \cite{hafner2019dream, hafner2020mastering}.

\subsection{Curiosity-Driven Exploration}
\label{curiosity}

An agent can be endowed with \textit{curiosity} by encouraging it to visit novel states, which can greatly improve exploration of the state space in the absence of dense extrinsic rewards. In large continuous state spaces, the prediction error of a learned dynamics model can act as a measure of novelty, and thus be leveraged as an intrinsic reward signal to create a curious agent \cite{burda2018large,li2019curiosity}. However, this approach can lead to the agent becoming `obsessed' with inherently unpredictable elements of the environment \cite{schmidhuber2010formal}. To balance this issue, Pathak et al. \cite{pathak2017curiosity} make predictions in a feature space learned by an inverse dynamics model, while Burda et al. \cite{burda2018exploration} employ a randomly initialised neural network to remove any stochasticity in the production of prediction targets. Intrinsic rewards have also be generated based on the disagreement between the next-state predictions of an ensemble of dynamics models \cite{pathak2019self, henaff2019explicit}. Such disagreement is more pronounced in regions of the state-space where training data is limited, thus acts as a good measure of novelty. Moreover, when repeatedly exposed to stochastic aspects of the environment, models will converge to predicting the mean and disagreement will diminish.

% Unlike standard prediction-error intrinsic rewards, the intrinsic rewards of \textcite{pathak2019self} are entirely self-generated, allowing them to be employed in imagination. \textcite{sekar2020planning} demonstrate this by learning an exploration policy in imagination then which covers the state-space highly efficiently when deployed in the real environment. This paper employs its ensemble of dynamics model not only to generate imaginary data but also to provide intrinsic rewards which directly encourage the collection of real data in areas where the model is uncertain.

\subsection{Model-Based Learning with Sparse Rewards}
Few proposed model-based methods have been well-equipped to handle difficult sparse-reward tasks. Indeed, Yang et al. \cite{yang2021mher} find that MBPO and MVE \cite{feinberg2018model} perform poorly in goal-based sparse-reward tasks. Multi-DEX \cite{kaushik2018multi}, a method tailored for sparse rewards, uses a curious model-based policy search algorithm which (as part of its multi-objective) seeks maximally novel state trajectories. Orthogonally, exploration policies have been learned efficiently in imagination via ensemble disagreement intrinsic rewards, aiding quick adaptation to downstream tasks \cite{shyam2019model,sekar2020planning}. However, these methods do not leverage the learning efficiencies that HER can provide in multi-goal tasks.

% Such policies cover the state space in a very efficient manner when deployed in the real environment, and build task-agnostic models that allow for quick adaptation to downstream tasks.
% Rather than operating on retrospective novelty, as seen in the methods of \ref{curiosity}, e
%  Such policies cover the state space in a very  efficient manner when deployed in the real environment, and build task-agnostic models that allow for quick adaptation to downstream tasks.

MHER \cite{yang2021mher} increases the efficiency of HER by generating imagined future achieved goals via a learned dynamics model, but this is only demonstrated in relatively trivial reaching tasks. Taking inspiration from HER, PlanGAN \cite{charlesworth2020plangan} trains an ensemble of generative models to generate trajectories that lead from the current state towards a specified goal, and use these trajectories for MPC. This improves efficiency versus HER, but entails a highly computationally expensive online planning procedure. 

Unlike these competing methods (\cite{yang2021mher,charlesworth2020plangan}), ours is demonstrated to be robust across the full set of benchmark Fetch tasks, providing roughly an order of magnitude increase in data-efficiency versus HER. This is achieved via a novel combination of curiosity-driven exploration, model-based learning, and HER, yielding an algorithm well suited to sparse-reward multi-goal tasks.

% Unlike these methods, ours is computationally efficient at inference, provides roughly an order of magnitude and is demonstrated to be robust across all tasks in the benchmark Fetch tasks.

% In I-HER...

\section{Background}

\subsection{Multi-goal Reinforcement Learning}
Our reinforcement learning agent interacts with a Markov decision process (MDP), defined by the tuple \((\mathcal{S},\mathcal{A},\mathcal{G},p,r,\gamma,\rho_0)\). $\mathcal{S}$, $\mathcal{A}$, and $\mathcal{G}$ are the state, action and goal spaces, respectively, and \(\gamma\in(0,1)\) is the discount factor. The state transition distribution is denoted as \(p(s'|s,a)\), the initial state distribution as \(\rho_0(s)\), and the reward function as \(r(s,g)\). The goal of the RL agent is to find the optimal policy $\pi^{*}$ that maximizes the expected sum of discounted rewards in this MDP: \(\pi^{*} = \operatorname*{argmax}_\pi \mathbb{E}_{\pi} [\sum_{t=0}^{\infty} \gamma^{t} r(s_t,g_t)]\).

The reward function is assumed to be known but the dynamics \(p(s'|s,a)\) are unknown. Model-based reinforcement learning constructs a model of the dynamics, \(\hat{f}_\theta(s'|s,a)\), using data collected from the MDP and supervised learning.

% \paragraph{Deep Deterministic Policy Gradients}
% Deep Deterministic Policy Gradients (DDPG) \cite{lillicrap2015continuous} is an off-policy model-free RL algorithm which maintains two neural networks: a policy (actor) \(\pi:\mathcal{S} \rightarrow \mathcal{A}\) and an action-value function (critic) $Q:\mathcal{S} \times \mathcal{A} \rightarrow \mathbb{R} $. The critic is trained using mini-batch gradient descent to minimise the loss \(\mathcal{L} = \mathbb{E}(Q(s_{t},a_{t})-y_{t})^2\), where $y_{t}=r_{t} + \gamma Q(s_{t+1}, \pi(s_{t+1}))$ and the tuples $(s_t,a_t,r_t,s_{t+1})$ are sampled from a replay buffer in which previously collected experience is stored. To stabilize this optimization procedure, the targets $y_{t}$ are computed using slowly updating target networks which are a polyak-averaged version of the main networks.

% The actor neural network is trained using mini-batch gradient descent to minimise the loss: $\mathcal{L}_a = -\mathbb{E}_s Q(s,\pi(s))$, where $s$ is sampled from the replay buffer and gradients are computed by backpropagating through the combined critic and actor networks.

\subsection{Hindsight Experience Replay (HER)}
HER can be used with any off-policy RL algorithm in multi-goal tasks, proving highly effective in these settings \cite{andrychowicz2017hindsight}. In these tasks, transition tuples collected in the MDP take the form of \((s_{t},a_{t},r_{t},s_{t+1},g)\), where $g$ is the goal of an episode, and the reward function is generally binary:
% \begin{center}
\begin{equation}
\label{sparse reward}
  r(s,g) =
    \begin{cases}
      0 & \text{if $g$ is satisfied by $s$}\\
     -1 & \text{otherwise}
    \end{cases}       
\end{equation}
% \end{center}

Now, if the goal is never reached in an episode in this MDP, all rewards will be negative and the collected transitions will provide minimal information for policy optimization. To solve this, HER employs a simple trick: a proportion of sampled transitions have their goal altered to $g’$, where $g’$ is a goal achieved later in the episode. The rewards of these altered transitions are then recalculated with respect to $g’$, leaving the altered transition tuples as $(s_{t},a_{t},r_t',s_{t+1},g')$. Even if the original episode was unsuccessful, the altered transitions will teach the agent how to achieve $g'$, thus accelerating its acquisition of skills.

% HER can be used with any off-policy RL algorithm and for simplicity we follow the original paper by employing Deep Deterministic Policy Gradients (DDPG) \cite{lillicrap2015continuous}.

\section{Imaginary Hindsight Experience Replay (I-HER)}

Excellent data-efficiency and the ability to learn from sparse rewards are appealing qualities for an RL algorithm being deployed in a robotic scenario. While HER has excelled in sparse-reward tasks where standard model-free RL algorithms have tended to fail \cite{andrychowicz2017hindsight,dai2020episodic}, it is apparent that its data efficiency could be improved through the use of a model-based component. In light of this, Imaginary Hindsight Experience Replay (I-HER) is now presented as a technique that incorporates imaginary/model-generated data into the HER learning procedure to improve its data efficiency. We follow the original paper by employing Deep Deterministic Policy Gradients (DDPG) \cite{lillicrap2015continuous} as the base off-policy RL algorithm. See Algorithm \ref{iher_algo} for the I-HER pseudo-code and refer to the appendix for comprehensive hyperparameter details.

% Many of the specific features of I-HER aim to minimize the adverse effects of learning from imaginary data generated by an inaccurate dynamics model, as discussed below.

\subsection{I-HER Overview}
I-HER consists of a continuous cycle of: (i) collecting real experience and updating the ensemble of dynamics models, then; (ii) generating imaginary experience and updating the policy. Collected real and imaginary MDP transitions are stored in their respective replay buffers, \(\mathcal{D}_{real}\) and \(\mathcal{D}_{imag}\). The dynamics model is trained on \(\mathcal{D}_{real}\). The RL policy is trained on data sampled from both \(\mathcal{D}_{real}\) and \(\mathcal{D}_{imag}\), combined into a single batch for the mini-batch gradient descent step. The sampling ratio here is based on the ratio of real to imaginary data collected so far: $ p(imag) = \frac{N(imag)}{N(imag) + N(real)} $, where $p(imag)$ is the proportion of sampled transitions that are imaginary, and $N(imag)$ and $N(real)$ are the respective number of imaginary and real transitions collected so far. This unbiased approach aims to avoid any overfitting that may occur from oversampling real experiences.

\subsection{Generating Imaginary Experience}
In I-HER, an ensemble of dynamics models \(\{\hat{f}_{\theta_{1}},...,\hat{f}_{\theta_{k}}\}\) is maintained. Following \cite{nagabandi2018neural,kurutach2018model}, each model is a feed-forward neural network  trained on data from \(\mathcal{D}_{real}\) to minimize the $L_{2}$ one-step prediction loss: $\|f(s_{t},a_{t};\theta)-(s_{t+1}-s_t)\|_{2}^{2}$, with normalized inputs and output targets. To generate imaginary experience, at each time step a model is sampled uniformly from the ensemble to predict $s_{t+1}$. This serves to minimise policy exploitation of any single models inaccuracies and thus improve policy robustness in the real environment \cite{kurutach2018model}. Diversity is maintained in the ensemble by varying model weight initializations and training input sequences \cite{kurutach2018model,clavera2018model}. Imaginary rollouts are of equal length to the real rollouts (50 steps in this case).

\subsection{Regenerating Imaginary Experience}
Off-policy RL algorithms learn from all previously collected experience maintained in the replay buffer. This presents an issue when learning from imaginary transitions: older imaginary transitions are likely more inaccurate than those generated by the latest dynamics model (which has been trained with the most data). Ideally, all imaginary experience should at least be consistent with the latest dynamics model, and I-HER ensures this is the case. Recognising that imaginary experience can be generated rapidly via batch parallelization, \(\mathcal{D}_{imag}\) is simply emptied each time the dynamic model is updated and refilled to its previous level with data generated by the up-to-date model\footnote{On an Intel Xeon Gold 6252 CPU, it takes less than 30 seconds to refill a full imaginary replay buffer with 1 million transitions.}. To ensure the buffer is still filled with a diverse set of experiences (an important factor for off-policy algorithms \cite{fedus2020revisiting,zhang2017deeper}), and to maintain learning stability, the new imaginary transitions are collected with the current and many older versions of the policy (see appendix for more details).

% When sampling from \(\mathcal{D}_{imag}\) to update the policy, these older transitions can have an adverse effect on performance. 

\subsection{Intrinsic Rewards}
In I-HER, the agent is provided with a useful intrinsic reward $r^{i}$, added to the extrinsic reward $r^{e}$ during DDPG updates in the same manner as \cite{li2019curiosity}, to give a total reward of: \(r_t=r_{t}^{e}+r_{t}^{i}\). Following \cite{pathak2019self}, this intrinsic reward is based on the variance between the ensemble predictions of $s_{t+1}$. Importantly, these ensemble-generated intrinsic rewards can be applied to imaginary data \cite{sekar2020planning}, and thus are applied to both real and imaginary transitions during policy updates. To weight the intrinsic rewards, and to ensure they do not overshadow the extrinsic rewards, they are scaled and clipped \cite{li2019curiosity}. Thus, the intrinsic reward is calculated as: \(r_{t}^{i}=\textrm{clip}(\nu\sigma_{t},0,\eta)\), where \(\sigma_{t}\) is the variance between ensemble predictions, $\nu$ is the scaling factor, and $\eta > 0$ is the maximum value $r_{t}$ can take.

This intrinsic reward signal serves two important purposes: (i) to strengthen the dynamics model by encouraging the collection of data in areas where it experiences epistemic uncertainty (uncertainty due to a lack of data), and (ii) to improve general exploration of the state-space.

\subsection{Distinguishing Real from Imaginary Experiences}
% The learned imaginary environment will never be a perfect replica of the real environment; especially when there is a deficit between model's capacity and the complexity of the dynamics. If highly precise control over the environment dynamics is required, even small model errors can hurt the performance of a policy learning solely from imaginary data. Naturally, training on a mixture of real and imaginary data will improve performance in this scenario, yet here the policy is being forced to generalise across both the real and imaginary environments. Unless it can differentiate between them, it will behave identically in each despite any differences.

When training in a imaginary environment, a model-based RL algorithm can overfit to inaccuracies in the imaginary dynamics at the expense of its real world performance. Now, to account for differences between a training environment and a target environment, meta-learning is commonly employed. For example, a recurrent policy can be learned whose memory allows it to discern the current dynamics and alter its behaviour appropriately \cite{andrychowicz2020learning}. Alternatively, a policy capable of quickly adapting to slight environmental differences in one gradient step can be learned \cite{clavera2018model}. A simpler solution is used in I-HER, albeit one that requires the use of \textit{some} real data to train the policy; the policy is told whether input observations are from the real or imaginary environment by appending to them a binary variable; a 1 if the observation is real, a 0 otherwise. This is done to allow the policy to detect and account for noticeable differences between the real and imaginary environments, and thus learn a set of behaviours fine-tuned for the real world.

% We hypothesise that by differentiating between reality and imagination, the policy can jointly optimize its performance in both

% This allows the policy to detect and account for differences between the real and imaginary environments, and thus learn a set of behaviours fine-tuned for the real world.  

% The policy is told whether its observations are from the real or imaginary environment by appending a binary variable '\texttt{env\char`_is\char`_real}' to its input: \texttt{env\char`_is\char`_real = 1} if the observation is from the real environment, 0 if the observation is imaginary.
% By differentiating between reality and imagination, the policy is allowed the opportunity to fine-tune behaviours for the real environment throughout its training.

\begin{algorithm}
\caption{Imaginary Hindsight Experience Replay (I-HER)}\label{iher_algo}
\begin{algorithmic}[1]

\State Initialize off-policy RL algorithm \(\mathbb{A}\) with policy \(\pi_\phi\)
\State Initialize ensemble of models {\(\{\hat{f}_{\theta_{1}},..,\hat{f}_{\theta_{k}}\}\)}
\State Initialize empty data sets \(\mathcal{D}_{real}\), \(\mathcal{D}_{imag}\), \(\mathcal{D}_{\pi}\)
\For{\textit{E} epochs}
    \State Collect \textit{R} real episodes with \(\pi_\phi\) and add to \(\mathcal{D}_{real}\)
    \State Train {\(\{\hat{f}_{\theta_{1}},..,\hat{f}_{\theta_{k}}\}\)} on \(\mathcal{D}_{real}\)
    \State Clear \(\mathcal{D}_{imag}\) and refill using \(\{{\hat{f}_{\theta_{1}},..,\hat{f}_{\theta_{k}}}\}\) and policies of \(\mathcal{D}_{\pi}\)
    \For{\textit{C} cycles}
        \State Collect \textit{I} imaginary episodes from \(\{{\hat{f}_{\theta_{1}},..,\hat{f}_{\theta_{k}}}\}\) with \(\pi_\phi\) and add to \(\mathcal{D}_{imag}\)
        \For{\textit{N} batches}
            \State Sample minibatch \(\textit{B}\) from \(\mathcal{D}_{real}\) and \(\mathcal{D}_{imag}\)
            \State Apply HER to \(\textit{B}\)
            \State Compute intrinsic rewards of \(\textit{B}\)
            \State Update \(\pi_\phi\) with \(\mathbb{A}\) and \(\textit{B}\)
        \EndFor
    \EndFor
    \State Add copy of \(\pi_\phi\) to \(\mathcal{D}_{\pi}\)
\EndFor

\end{algorithmic}
\end{algorithm}

\section{Experiments}

\subsection{Environments}

\begin{figure}[h]
\centering
    \includegraphics[width=0.48\textwidth]{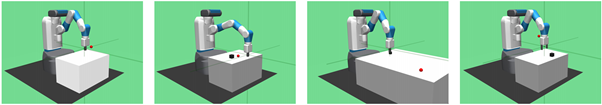}
    \caption{The OpenAI Gym Fetch Robotics tasks: Reach, Push, Slide, and PickAndPlace \cite{plappert2018multi}.}
\label{FetchEnvs}
\end{figure}

I-HER is evaluated in the OpenAI Gym Fetch Robotics environments \cite{plappert2018multi}, which consist of four diverse and challenging continuous control tasks: Reach, Push, PickAndPlace, and Slide (see Fig. \ref{FetchEnvs}). Rewards in these multi-goal tasks are sparse and binary (as in Eq. (\ref{sparse reward})). Observations include relevant state information.

In these tasks, HER is known to outperform standard model-free algorithms, which struggle due to the sparse reward function \cite{andrychowicz2017hindsight,dai2020episodic}. Although there have been several iterations on the original HER (e.g. \cite{he2020soft,fang2019curriculum}), these have generally provided incremental improvements in performance, and could be integrated into the I-HER algorithm if desired. As such, we regard HER as the state-of-the-art model-free method in the benchmark Fetch tasks for the purposes of comparison to I-HER.

% The sparse reward function makes exploration very difficult, and standard model-free RL algorithms often completely fail to learn in these tasks \cite{andrychowicz2017hindsight,dai2020episodic}. Through its implicit curriculum learning, HER greatly simplifies exploration and solves these tasks relatively efficiently. Without HER, curiosity-based methods \cite{li2019curiosity} and a self-imitation learning method \cite{dai2020episodic} have solved the tasks but did so requiring significantly more data. Although there have been several iterations on the original HER \cite{andrychowicz2017hindsight}, these additions generally provide incremental improvements in performance and can be integrated into the I-HER method if desired. As such, for the purposes of comparison, we regard HER as the state-of-the-art method in the Fetch tasks. 

\subsection{Benchmark Results}

\begin{figure}[h]
\centering
\includegraphics[width=0.48\textwidth]{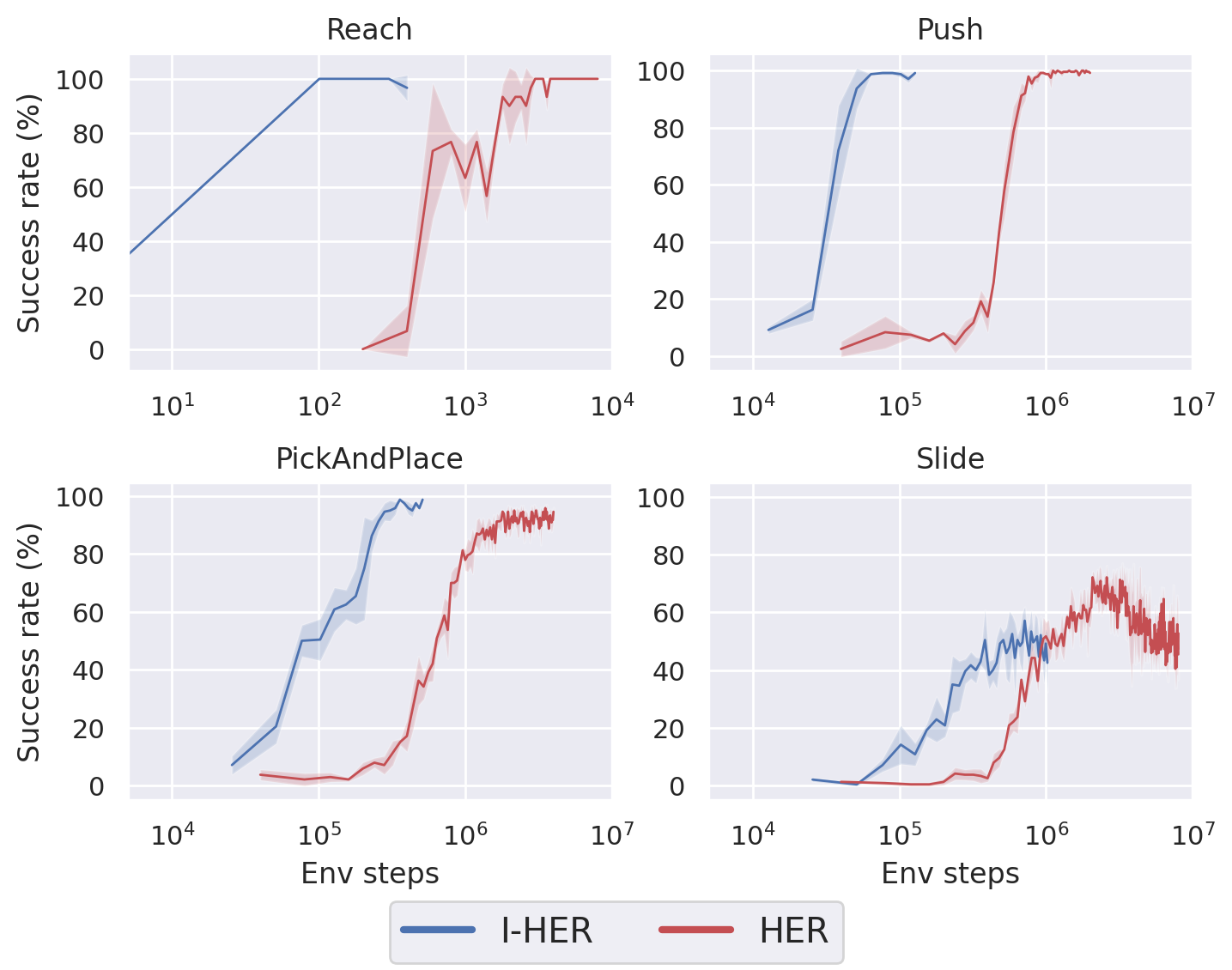}
\caption{I-HER's performance is compared to HER, the state-of-the-art in the Fetch tasks. The x-axis, in \textit{log-scale}, denotes the number of time steps of real environment data used. The y-axis denotes task success rate in the real environment. These figures demonstrate that I-HER provides substantial improvements in sample complexity versus HER.\protect\footnotemark} 
\label{I-HER}
\end{figure}

The performance of I-HER is now compared to HER\footnote{All presented experiments were performed over 3 random seeds}. Both methods were implemented with DDPG as the base RL algorithm. To enable fair comparison, shared hyperparameters were maintained at equal values, largely based on those employed by \cite{plappert2018multi}. The results of this comparative evaluation are shown in Fig. \ref{I-HER}. 

The model-based I-HER matches the asymptotic performance of the model-free HER in all tasks except Slide, requiring on average an order of magnitude less real data to learn. Roughly 30$\times$, 15$\times$, 7$\times$, and 4$\times$ less data are required in Reach, Push, PickAndPlace, and Slide, respectively (notably, gains diminish as task difficulty increases).  To solve the non-trivial Push task, I-HER requires 40 minutes of experience versus the 10 hours required by HER. Interestingly, I-HER achieves a slightly higher asymptotic performance in the PickAndPlace task, perhaps due to the improved exploration encouraged by its intrinsic rewards.

Turning now to the reported results of competing model-based methods. Of the Fetch tasks, MHER \cite{yang2021mher} is only tested in the trivial Reach but does not match our 30 times improvement in data-efficiency versus HER. PlanGAN \cite{charlesworth2020plangan} displays similar improvements in PickAndPlace but do not match our 15 times improvement in efficiency in Push.

\begin{figure}[h]
\centering
\includegraphics[width=0.48\textwidth]{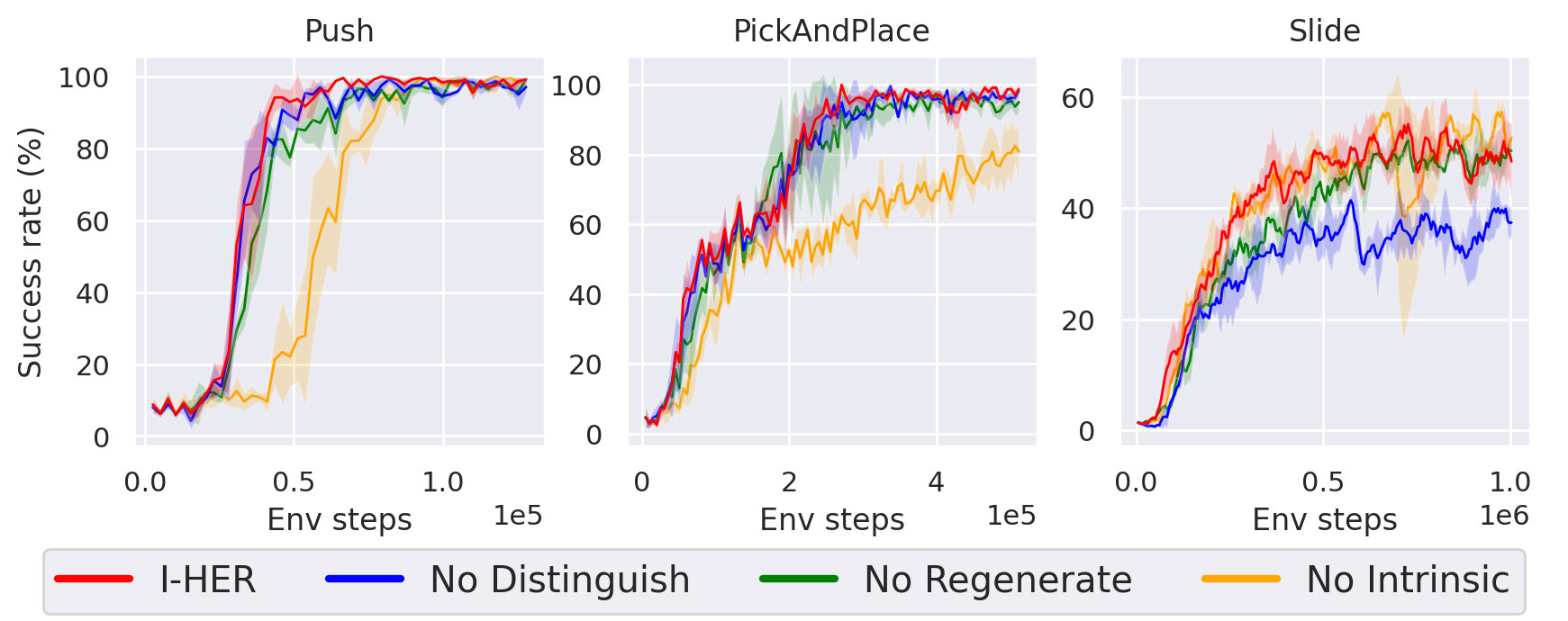}
\caption{Contributions of I-HER's components. The full I-HER method (red) is compared to: I-HER without the distinguishing of real and imaginary experiences (blue), without regenerating imaginary data (green), and without intrinsic rewards (yellow). The Slide results are averaged over five epochs for improved readability.}
\label{compare}
\end{figure}

\begin{figure}[h]
\centering
\includegraphics[width=0.48\textwidth]{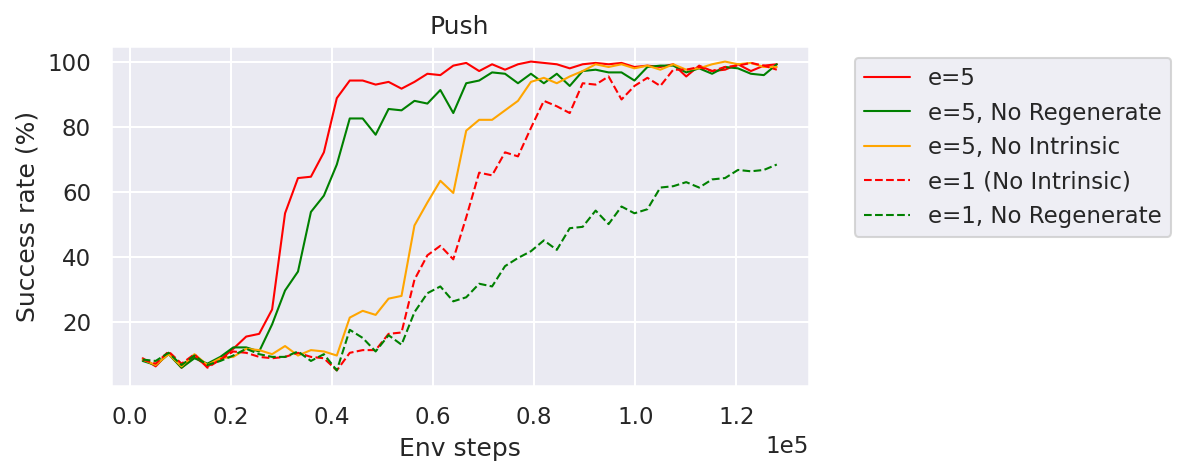}
\caption{Analysis of the effect of \textit{e = ensemble size} in the Push task. Performance is highest with an ensemble of five models as useful intrinsic rewards are generated and the policy cannot easily overfit to model inaccuracies. With an `ensemble' of one model, however, no intrinsic rewards are generated (ensemble variance is always zero), and the policy significantly overfits to model inaccuracies when old imaginary experiences are not regenerated. Note, variance in results is omitted for improved readability.}
\label{ensemble}
\end{figure}

\subsection{Ablation Study}

% Figs. \ref{compare} and \ref{ensemble} assess the contributions of the components of I-HER to its final performance, the results of which are now discussed.

\paragraph{Intrinsic rewards} The intrinsic rewards prove very useful in Push and PickAndPlace (see Fig. \ref{compare}), serving to encourage to interaction with the cube, whose dynamics are complex and difficult to model. Without curiosity, the agent struggles to learn to pickup the cube in PickAndPlace. Here, when faced with a goal it was unable to reach, it was observed that the curious agent would continue to `play' with the cube while the non-curious agent would remain relatively inactive. This `play' was crucial for the agent to quickly learn to pick up the cube and move it to goals above the table.

\paragraph{Distinguishing between reality and imagination}

\footnotetext{Unlike in Figs \ref{compare} and \ref{ensemble}, the I-HER results in Fig \ref{I-HER} do not include \textit{within} epoch evaluations, hence their smoother appearance.}

Contrary to the intrinsic rewards, distinguishing real and imaginary experiences only provides notable benefits in Slide (see Fig. \ref{compare}). This can be explained as follows. In Push and PickAndPlace, the bottleneck in learning the task is exploration; once the agent has adequately explored the cube, the success rate quickly converges to 100\%. In Slide, however, highly precise control is required (as demonstrated by the inability of I-HER or HER to reach an 100\% success rate in Fig. \ref{I-HER}), and this becomes the bottleneck rather than exploration; the Slide policy is still being fine-tuned long after it has learned to interact with the puck.

Since highly precise control over the dynamics is required, small modelling errors can hurt policy performance significantly. Distinguishing real and imaginary data allows the policy to account for any noticeable differences in the real and imaginary dynamics, and thus learn a set of behaviors fine-tuned for the real world. This helps minimise the adverse effect of model errors, providing a significant boost in I-HER's final Slide performance.

\paragraph{Regenerating imaginary experiences} The results of Fig. \ref{compare} suggest that the regeneration of imaginary experiences only provides minor benefits in the final version of I-HER (with an ensemble of five models). However, an analysis of the effect of the ensemble (see Figure \ref{ensemble}) demonstrates that this regeneration step is highly beneficial when only a single dynamics model is employed.

With a single model, model inaccuracies can be easily exploited in the imaginary environment. For example, at one stage the policy was observed capable of `magically' moving the imaginary cube without touching it, due to a poor dynamics model. Such inaccuracies are generally corrected once the policy collects further data in real environment, however, if the inaccurate imaginary data remains in the buffer, the policy will continue to try to exploit it. Thus, regenerating the imaginary experience in the buffer is crucial when only a single dynamics model is employed. With an ensemble of five models, the policy cannot overfit to model inaccuracies as easily and maintaining older imaginary experiences in the buffer is not as detrimental to performance.

% \section{Discussion}

\section{Challenges in Model-Based RL}

The diversity of the Fetch tasks expose two modes in which model-based RL methods can struggle: (a) where significant exploratory barriers must be passed; (b) where highly precise control over complex dynamics is necessary. These are now discussed in the context of the I-HER results.

\paragraph{Exploration}
In model-based RL, the policy and model are co-dependant. The model depends on the policy to provide it with data, while the policy requires an adequately accurate model to aid its learning of the task. When exploration is difficult, this codependency becomes an issue. This is evident in PickAndPlace, where learning to pick up the cube is a major exploratory barrier. While the intrinsic rewards are highly beneficial here, I-HER still struggles to breach this barrier; as demonstrated by the learning plateaus around the 50-65\% success rate in the PickAndPlace results of Fig. \ref{compare} (half of the goals are on the table while the remainder are above). Analysis found that before and during this plateau it was impossible to properly pick up the cube in the imaginary environment; the model had not yet seen any cube-picking data, so its predictions were biased towards the cube falling back to the table. Hence, the policy struggled to learn to pick up the cube as it was impossible to do so in imaginary environment at this stage. This in turn prevented the policy from providing the model with the cube-picking data it needed to improve its predictions, which in turn continued to hurt the policies ability to learn cube-picking, and so on. This unfortunate cycle that was difficult to break without the use of intrinsic rewards.

% resulting in the policy getting stuck in a local minimum that was difficult to
% Assuming the model has the capacity to represent the dynamics accurately, the onus is on the policy to explore the environment effectively and provide the model with task relevant data, else their codependent relationship will fail.

\paragraph{Precision}
Slide requires highly precise control over complex dynamics. The `slide' must be perfectly weighted and directed, and once the puck moves out of reach any misjudgements cannot be corrected. Thus, small biases in the imaginary environment can hurt policy performance substantially. For example, if the imagined friction is slightly too low, the policy will tend to leave its shots too short in the real environment. Ultimately, the requirement for high precision in both the dynamical modelling and the control policy leads to suboptimal I-HER results in Slide.

% Distinguish method a bit crude and could be improved

These exploration and precision issues are resolved to some extent by I-HER's intrinsic rewards and distinguishing of real and imaginary experiences, respectively. However, there is room for improvement. An exploration mechanism better able to identify completely unvisited areas of the state space would be beneficial; in PickAndPlace the policy can consistently receive easy intrinsic rewards by aggressively fumbling the cube on the table, thus reducing its incentive to explore picking up the cube. Learning a separate Q-function for intrinsic and extrinsic rewards, similar to \cite{lanier2019curiosity}, could be beneficial. Enhancing model capacity to improve predictions would likely help with both issues. This could potentially be achieved by increasing the model size, using probabilistic dynamics models \cite{chua2018deep} (particularly useful if dealing with the stochasticity of the real world), or by introducing a multi-step prediction objective \cite{hafner2019learning}.

% In I-HER, incorporating distinguished real and imaginary data into policy updates helped reduce the negative effects of model biases, however, the maximum success rates achieved by HER were not reached.

%That HER only reaches an maximum success rate of 80\% in this task is evidence of the precision required. 

\section{Conclusion}

Solving difficult robotic tasks via reinforcement learning can require the collection of excessive amounts of data, while designing and implementing a suitably shaped reward function can require expert knowledge and complex engineering. To address these issues, this paper proposes Imaginary Hindsight Experience Replay (I-HER); a model-based method specifically tailored for sparse-reward multi-goal tasks. Empirical results demonstrate that this method significantly reduces sample complexity versus the state-of-the-art model-free method in challenging robotic tasks, matching its asymptotic performance in all but one task.
 
An important direction for future work will be to extend this approach to visual observations; acquiring state observations in the real world is challenging and does not scale well to increasingly unstructured settings. This would require the adaption of the concept of HER to handle image observations, as attempted in \cite{nair2018visual,sahni2019addressing}. Finally, the benefits of I-HER should be verified via its deployment in a real-world robotics scenarios, further testing the limits of model-based reinforcement learning.

\section*{APPENDIX}
% - Like in (Kurutach et al, 2018; Clavera et al, 2018), the reward function used in imagination is not learned but has been provided by the real environment.

% - FetchSlide: TEST WHERE SHOTS ARE MISSED, TEST WHETHER IMAG POLICY PERFORMS WORSE IN REAL ENV THAN REAL POLICY, TEST DIFFER REAL IMAG POLICIES IN GENERAL)

% - (APPENDIX?:) Interestingly, I-HER learns a more conservative policy than HER - rather than confidently hitting the puck first time, it will often tap it as close to the target as possible before finally hitting it out of its reach. This behaviour is likely induced by the variance of the imaginary environment - each model of the ensemble is slightly different and models are updated regularly.

% \subsection{HER Details}
% In the HER experiments, 1 MPI worker is used in Reach and 8 are used in the remaining tasks\footnote{Our HER implementation is taken from \url{https://github.com/TianhongDai/hindsight-experience-replay}}. Besides this, all relevant hyperparameters are kept equivalent to \cite{plappert2018multi}.

% \subsection{I-HER Details}

\paragraph{Dynamics model} An ensemble of 5 models is maintained. Each model is a feed-forward neural network with 2 hidden layers of size 512 and ReLU nonlinearities. Models are trained with the Adam optimizer \cite{kingma2014adam} with a learning rate of 0.001 and batch size of 512 (each MPI worker samples its own batch of size 512 and gradients are averaged for the update step). Each epoch, each model is trained for \textit{K} steps, jumpstarted off the previous epochs model. In the first epoch, the model is trained for 5\textit{K} steps.

% In general, it was found that increasing \textit{K} improved performance, however, using very high values of \textit{K} can be time-consuming (especially as no GPU was used in this papers experiments).
% \begin{equation}
%     p_{i} = \frac{w_{i}}{\sum{w_{i}}}
% \end{equation}

% \begin{equation}  
%     w_{i} = \frac{E_{i}}{E}(b - 1)
% \end{equation}
% , meaning transitions from the latest epoch are twice as likely to be sampled as those from the first epoch. This allows the model to focus on improving its predictions for newer transitions rather than older ones which it has seen many times previously. 

To attempt to improve the efficiency of the update steps, a bias is introduced when sampling from \(\mathcal{D}_{real}\) to update the dynamics models. Each transition $i$ is sampled with probability $p(i) = \frac{w(i)}{\sum{w(i)}}$, where $w(i) = \frac{E(i)}{E}(b - 1)$, $E(i)$ is the epoch in which $i$ was collected, $E$ is the current epoch, and $b \geqslant 1$ decides how much to bias towards newer transitions. In all experiments, $b=2$ is used. 

\paragraph{Hyperparameters}
The following hyperparameters are equivalent across all tasks: \(\mathcal{D}_{real}\), \(\mathcal{D}_{imag}\) buffer sizes = $10^6$; imaginary rollouts per MPI worker (\textit{I}) = 2; batches per cycle (\textit{N}) = 40; intrinsic reward scale ($\nu$) = 0.5; intrinsic reward clip ($\eta$) = 0.8. The following are different across $\langle$Reach, Push, PickAndPlace, Slide$\rangle$; MPI workers = $\langle$1, 8, 8, 8$\rangle$; dynamics update steps (\textit{K}) = $\langle$50, 1250, 2500, 2500$\rangle$; real rollouts per MPI worker (\textit{R}) = $\langle$2, 32, 64, 64$\rangle$; cycles (\textit{C}) = $\langle$50, 250, 250, 250$\rangle$.

%\paragraph{I-HER}

% \begin{table*}[t]
% \caption{An Example of a Table}
% \begin{center}
% \begin{tabular}{ |l|c|c|c|c| } 
%  \hline
%   & Reach & Push & PickAndPlace & Slide \\
%  \hline
%  MPI workers & 1 & \multicolumn{3}{|c|}{8}\\
%  \hline
%  \(\mathcal{D}_{real}\), \(\mathcal{D}_{imag}\) buffer sizes & \multicolumn{4}{|c|}{$10^6$} \\
%  \hline
%  Dynamics update steps (\textit{K}) & 100 & 1250 & \multicolumn{2}{|c|}{2500}\\
%  \hline
%  Real rollouts per MPI worker (\textit{R}) & 5 & 32 & \multicolumn{2}{|c|}{64} \\
%  \hline
%  Cycles (\textit{C}) & \multicolumn{4}{|c|}{250} \\
%  \hline
%  Imaginary rollouts per MPI worker (\textit{I}) & \multicolumn{4}{|c|}{2} \\
%  \hline
%  Batches per cycle (\textit{N}) & \multicolumn{4}{|c|}{40} \\
%  \hline
%  Intrinsic reward scale ($\nu$) & \multicolumn{4}{|c|}{0.5} \\
%  \hline
%  Intrinsic reward clip ($\eta$) & \multicolumn{4}{|c|}{0.8} \\
%  \hline
% \end{tabular}
% \end{center}
% \end{table*}

\paragraph{Refilling imaginary replay buffer (line 7 of Algorithm \ref{iher_algo})} Iterating through the policies stored in \(\mathcal{D}_{\pi}\) from newest to oldest, each policy collects $C \times I$ imaginary rollouts which are added to \(\mathcal{D}_{imag}\). This process stops once \(\mathcal{D}_{imag}\) is refilled to its previous level. A copy of \(\pi_\phi\) is added to \(\mathcal{D}_{\pi}\) every 50 cycles (\textit{C}).
% The hyperparameter $b_{\pi} \geqslant 1$ controls how 'on-policy' the data being collected is; at greater values the rollouts are collected by the more recent versions of the policy. At $b_{\pi} = 1$ the rollouts are collected equally by all saved versions of the policy. Since increasing $b_{\pi}$ did not show any clear benefits, the 'default' value of $b_{\pi} = 1$ is used in all experiments.

% %%%%%%%%%%%%%%%%%%%%%%%%%
% \begin{algorithm}
% \caption{Refilling imaginary buffer (line 7 of Algorithm 1)}\label{euclid}
% \begin{algorithmic}[1]

% \State Clear \(\mathcal{D}_{imag}\)
% \For{$\pi$ in \(\mathcal{D}_{\pi}\)}
%     \State Collect ($C \times I$) imaginary rollouts with $\pi$ and add to \(\mathcal{D}_{imag}\)
%     \If{\(\mathcal{D}_{imag}\) is refilled to previous level}
%         \State \textbf{return}
%     \EndIf
% \EndFor

% \end{algorithmic}
% \end{algorithm}
% %%%%%%%%%%%%%%%%%%%%%%%%%

\paragraph{Distinguishing between real and imaginary data}
During updates, the policy is always truthfully told whether a transition is real or imaginary (via the binary variable \textit{env\_is\_real}). However, when collecting experience, 10\% of real rollouts are collected with \textit{env\_is\_real = 0} as input, and 10\% of imaginary rollouts are collected with \textit{env\_is\_real = 1} as input. This mixing is to (i) encourage the real and imaginary behaviours to learn from each other (beyond their shared weights), and (ii) to ensure exploiting imaginary behaviours are corrected by testing them in the real environment.

\paragraph{HER experiments}
1 MPI worker is used in Reach and 8 are used in the remaining tasks\footnote{Our HER implementation is taken from \url{https://github.com/TianhongDai/hindsight-experience-replay}}. Besides this, all relevant hyperparameters are kept equivalent to \cite{plappert2018multi}.

\AtNextBibliography{\small}

% If biblatex
% \bibliographystyle{numeric}
\printbibliography

@article{mnih2015human,
  title={Human-level control through deep reinforcement learning},
  author={Mnih, Volodymyr and Kavukcuoglu, Koray and Silver, David and Rusu, Andrei A and Veness, Joel and Bellemare, Marc G and Graves, Alex and Riedmiller, Martin and Fidjeland, Andreas K and Ostrovski, Georg and others},
  journal={nature},
  volume={518},
  number={7540},
  pages={529--533},
  year={2015},
  publisher={Nature Publishing Group}
}

@article{berner2019dota,
  title={Dota 2 with large scale deep reinforcement learning},
  author={Berner, Christopher and Brockman, Greg and Chan, Brooke and Cheung, Vicki and D{\k{e}}biak, Przemys{\l}aw and Dennison, Christy and Farhi, David and Fischer, Quirin and Hashme, Shariq and Hesse, Chris and others},
  journal={arXiv preprint arXiv:1912.06680},
  year={2019}
}

@article{janner2019trust,
  title={When to trust your model: Model-based policy optimization},
  author={Janner, Michael and Fu, Justin and Zhang, Marvin and Levine, Sergey},
  journal={arXiv preprint arXiv:1906.08253},
  year={2019}
}

@article{hafner2020mastering,
  title={Mastering atari with discrete world models},
  author={Hafner, Danijar and Lillicrap, Timothy and Norouzi, Mohammad and Ba, Jimmy},
  journal={arXiv preprint arXiv:2010.02193},
  year={2020}
}

@article{schrittwieser2020mastering,
  title={Mastering atari, go, chess and shogi by planning with a learned model},
  author={Schrittwieser, Julian and Antonoglou, Ioannis and Hubert, Thomas and Simonyan, Karen and Sifre, Laurent and Schmitt, Simon and Guez, Arthur and Lockhart, Edward and Hassabis, Demis and Graepel, Thore and others},
  journal={Nature},
  volume={588},
  number={7839},
  pages={604--609},
  year={2020},
  publisher={Nature Publishing Group}
}

@inproceedings{pathak2017curiosity,
  title={Curiosity-driven exploration by self-supervised prediction},
  author={Pathak, Deepak and Agrawal, Pulkit and Efros, Alexei A and Darrell, Trevor},
  booktitle={International Conference on Machine Learning},
  pages={2778--2787},
  year={2017},
  organization={PMLR}
}

@article{burda2018exploration,
  title={Exploration by random network distillation},
  author={Burda, Yuri and Edwards, Harrison and Storkey, Amos and Klimov, Oleg},
  journal={arXiv preprint arXiv:1810.12894},
  year={2018}
}

@article{andrychowicz2017hindsight,
  title={Hindsight experience replay},
  author={Andrychowicz, Marcin and Wolski, Filip and Ray, Alex and Schneider, Jonas and Fong, Rachel and Welinder, Peter and McGrew, Bob and Tobin, Josh and Abbeel, Pieter and Zaremba, Wojciech},
  journal={arXiv preprint arXiv:1707.01495},
  year={2017}
}

@article{kurutach2018model,
  title={Model-ensemble trust-region policy optimization},
  author={Kurutach, Thanard and Clavera, Ignasi and Duan, Yan and Tamar, Aviv and Abbeel, Pieter},
  journal={arXiv preprint arXiv:1802.10592},
  year={2018}
}

@inproceedings{pathak2019self,
  title={Self-supervised exploration via disagreement},
  author={Pathak, Deepak and Gandhi, Dhiraj and Gupta, Abhinav},
  booktitle={International Conference on Machine Learning},
  pages={5062--5071},
  year={2019},
  organization={PMLR}
}

@inproceedings{nagabandi2018neural,
  title={Neural network dynamics for model-based deep reinforcement learning with model-free fine-tuning},
  author={Nagabandi, Anusha and Kahn, Gregory and Fearing, Ronald S and Levine, Sergey},
  booktitle={2018 IEEE International Conference on Robotics and Automation (ICRA)},
  pages={7559--7566},
  year={2018},
  organization={IEEE}
}

@inproceedings{nagabandi2020deep,
  title={Deep dynamics models for learning dexterous manipulation},
  author={Nagabandi, Anusha and Konolige, Kurt and Levine, Sergey and Kumar, Vikash},
  booktitle={Conference on Robot Learning},
  pages={1101--1112},
  year={2020},
  organization={PMLR}
}

@inproceedings{hafner2019learning,
  title={Learning latent dynamics for planning from pixels},
  author={Hafner, Danijar and Lillicrap, Timothy and Fischer, Ian and Villegas, Ruben and Ha, David and Lee, Honglak and Davidson, James},
  booktitle={International Conference on Machine Learning},
  pages={2555--2565},
  year={2019},
  organization={PMLR}
}

@article{chua2018deep,
  title={Deep reinforcement learning in a handful of trials using probabilistic dynamics models},
  author={Chua, Kurtland and Calandra, Roberto and McAllister, Rowan and Levine, Sergey},
  journal={arXiv preprint arXiv:1805.12114},
  year={2018}
}

@article{feinberg2018model,
  title={Model-based value estimation for efficient model-free reinforcement learning},
  author={Feinberg, Vladimir and Wan, Alvin and Stoica, Ion and Jordan, Michael I and Gonzalez, Joseph E and Levine, Sergey},
  journal={arXiv preprint arXiv:1803.00101},
  year={2018}
}

@article{buckman2018sample,
  title={Sample-efficient reinforcement learning with stochastic ensemble value expansion},
  author={Buckman, Jacob and Hafner, Danijar and Tucker, George and Brevdo, Eugene and Lee, Honglak},
  journal={arXiv preprint arXiv:1807.01675},
  year={2018}
}

@incollection{sutton1990integrated,
  title={Integrated architectures for learning, planning, and reacting based on approximating dynamic programming},
  author={Sutton, Richard S},
  booktitle={Machine learning proceedings 1990},
  pages={216--224},
  year={1990},
  publisher={Elsevier}
}

@inproceedings{gu2016continuous,
  title={Continuous deep q-learning with model-based acceleration},
  author={Gu, Shixiang and Lillicrap, Timothy and Sutskever, Ilya and Levine, Sergey},
  booktitle={International Conference on Machine Learning},
  pages={2829--2838},
  year={2016},
  organization={PMLR}
}

@inproceedings{clavera2018model,
  title={Model-based reinforcement learning via meta-policy optimization},
  author={Clavera, Ignasi and Rothfuss, Jonas and Schulman, John and Fujita, Yasuhiro and Asfour, Tamim and Abbeel, Pieter},
  booktitle={Conference on Robot Learning},
  pages={617--629},
  year={2018},
  organization={PMLR}
}

@article{luo2018algorithmic,
  title={Algorithmic framework for model-based deep reinforcement learning with theoretical guarantees},
  author={Luo, Yuping and Xu, Huazhe and Li, Yuanzhi and Tian, Yuandong and Darrell, Trevor and Ma, Tengyu},
  journal={arXiv preprint arXiv:1807.03858},
  year={2018}
}

@article{kaiser2019model,
  title={Model-based reinforcement learning for atari},
  author={Kaiser, Lukasz and Babaeizadeh, Mohammad and Milos, Piotr and Osinski, Blazej and Campbell, Roy H and Czechowski, Konrad and Erhan, Dumitru and Finn, Chelsea and Kozakowski, Piotr and Levine, Sergey and others},
  journal={arXiv preprint arXiv:1903.00374},
  year={2019}
}

@article{ha2018world,
  title={World models},
  author={Ha, David and Schmidhuber, J{\"u}rgen},
  journal={arXiv preprint arXiv:1803.10122},
  year={2018}
}

@article{burda2018large,
  title={Large-scale study of curiosity-driven learning},
  author={Burda, Yuri and Edwards, Harri and Pathak, Deepak and Storkey, Amos and Darrell, Trevor and Efros, Alexei A},
  journal={arXiv preprint arXiv:1808.04355},
  year={2018}
}

@article{schmidhuber2010formal,
  title={Formal theory of creativity, fun, and intrinsic motivation (1990--2010)},
  author={Schmidhuber, J{\"u}rgen},
  journal={IEEE Transactions on Autonomous Mental Development},
  volume={2},
  number={3},
  pages={230--247},
  year={2010},
  publisher={Ieee}
}

@article{henaff2019explicit,
  title={Explicit explore-exploit algorithms in continuous state spaces},
  author={Henaff, Mikael},
  journal={arXiv preprint arXiv:1911.00617},
  year={2019}
}

@inproceedings{kaushik2018multi,
  title={Multi-objective model-based policy search for data-efficient learning with sparse rewards},
  author={Kaushik, Rituraj and Chatzilygeroudis, Konstantinos and Mouret, Jean-Baptiste},
  booktitle={Conference on Robot Learning},
  pages={839--855},
  year={2018},
  organization={PMLR}
}

@inproceedings{sekar2020planning,
  title={Planning to explore via self-supervised world models},
  author={Sekar, Ramanan and Rybkin, Oleh and Daniilidis, Kostas and Abbeel, Pieter and Hafner, Danijar and Pathak, Deepak},
  booktitle={International Conference on Machine Learning},
  pages={8583--8592},
  year={2020},
  organization={PMLR}
}

@inproceedings{shyam2019model,
  title={Model-based active exploration},
  author={Shyam, Pranav and Ja{\'s}kowski, Wojciech and Gomez, Faustino},
  booktitle={International conference on machine learning},
  pages={5779--5788},
  year={2019},
  organization={PMLR}
}

@article{yang2021mher,
  title={MHER: Model-based Hindsight Experience Replay},
  author={Yang, Rui and Fang, Meng and Han, Lei and Du, Yali and Luo, Feng and Li, Xiu},
  journal={arXiv preprint arXiv:2107.00306},
  year={2021}
}

@article{charlesworth2020plangan,
  title={Plangan: Model-based planning with sparse rewards and multiple goals},
  author={Charlesworth, Henry and Montana, Giovanni},
  journal={arXiv preprint arXiv:2006.00900},
  year={2020}
}

@article{lillicrap2015continuous,
  title={Continuous control with deep reinforcement learning},
  author={Lillicrap, Timothy P and Hunt, Jonathan J and Pritzel, Alexander and Heess, Nicolas and Erez, Tom and Tassa, Yuval and Silver, David and Wierstra, Daan},
  journal={arXiv preprint arXiv:1509.02971},
  year={2015}
}

@article{dai2020episodic,
  title={Episodic Self-Imitation Learning with Hindsight},
  author={Dai, Tianhong and Liu, Hengyan and Anthony Bharath, Anil},
  journal={Electronics},
  volume={9},
  number={10},
  pages={1742},
  year={2020},
  publisher={Multidisciplinary Digital Publishing Institute}
}

@inproceedings{fedus2020revisiting,
  title={Revisiting fundamentals of experience replay},
  author={Fedus, William and Ramachandran, Prajit and Agarwal, Rishabh and Bengio, Yoshua and Larochelle, Hugo and Rowland, Mark and Dabney, Will},
  booktitle={International Conference on Machine Learning},
  pages={3061--3071},
  year={2020},
  organization={PMLR}
}

@article{zhang2017deeper,
  title={A deeper look at experience replay},
  author={Zhang, Shangtong and Sutton, Richard S},
  journal={arXiv preprint arXiv:1712.01275},
  year={2017}
}

@article{andrychowicz2020learning,
  title={Learning dexterous in-hand manipulation},
  author={Andrychowicz, OpenAI: Marcin and Baker, Bowen and Chociej, Maciek and Jozefowicz, Rafal and McGrew, Bob and Pachocki, Jakub and Petron, Arthur and Plappert, Matthias and Powell, Glenn and Ray, Alex and others},
  journal={The International Journal of Robotics Research},
  volume={39},
  number={1},
  pages={3--20},
  year={2020},
  publisher={SAGE Publications Sage UK: London, England}
}

@inproceedings{li2019curiosity,
  title={Curiosity-driven exploration for off-policy reinforcement learning methods},
  author={Li, Boyao and Lu, Tao and Li, Jiayi and Lu, Ning and Cai, Yinghao and Wang, Shuo},
  booktitle={2019 IEEE International Conference on Robotics and Biomimetics (ROBIO)},
  pages={1109--1114},
  year={2019},
  organization={IEEE}
}

@article{plappert2018multi,
  title={Multi-goal reinforcement learning: Challenging robotics environments and request for research},
  author={Plappert, Matthias and Andrychowicz, Marcin and Ray, Alex and McGrew, Bob and Baker, Bowen and Powell, Glenn and Schneider, Jonas and Tobin, Josh and Chociej, Maciek and Welinder, Peter and others},
  journal={arXiv preprint arXiv:1802.09464},
  year={2018}
}

@article{nair2018visual,
  title={Visual reinforcement learning with imagined goals},
  author={Nair, Ashvin and Pong, Vitchyr and Dalal, Murtaza and Bahl, Shikhar and Lin, Steven and Levine, Sergey},
  journal={arXiv preprint arXiv:1807.04742},
  year={2018}
}

@article{sahni2019addressing,
  title={Addressing Sample Complexity in Visual Tasks Using HER and Hallucinatory GANs},
  author={Sahni, Himanshu and Buckley, Toby and Abbeel, Pieter and Kuzovkin, Ilya},
  journal={arXiv preprint arXiv:1901.11529},
  year={2019}
}

@inproceedings{kalweit2017uncertainty,
  title={Uncertainty-driven imagination for continuous deep reinforcement learning},
  author={Kalweit, Gabriel and Boedecker, Joschka},
  booktitle={Conference on Robot Learning},
  pages={195--206},
  year={2017},
  organization={PMLR}
}

@article{popov2017data,
  title={Data-efficient deep reinforcement learning for dexterous manipulation},
  author={Popov, Ivaylo and Heess, Nicolas and Lillicrap, Timothy and Hafner, Roland and Barth-Maron, Gabriel and Vecerik, Matej and Lampe, Thomas and Tassa, Yuval and Erez, Tom and Riedmiller, Martin},
  journal={arXiv preprint arXiv:1704.03073},
  year={2017}
}

@article{brockman2016openai,
  title={Openai gym},
  author={Brockman, Greg and Cheung, Vicki and Pettersson, Ludwig and Schneider, Jonas and Schulman, John and Tang, Jie and Zaremba, Wojciech},
  journal={arXiv preprint arXiv:1606.01540},
  year={2016}
}

@book{lanier2019curiosity,
  title={Curiosity-driven multi-criteria hindsight experience replay},
  author={Lanier, John Banister},
  year={2019},
  publisher={University of California, Irvine}
}

@article{he2020soft,
  title={Soft hindsight experience replay},
  author={He, Qiwei and Zhuang, Liansheng and Li, Houqiang},
  journal={arXiv preprint arXiv:2002.02089},
  year={2020}
}

@article{fang2019curriculum,
  title={Curriculum-guided hindsight experience replay},
  author={Fang, Meng and Zhou, Tianyi and Du, Yali and Han, Lei and Zhang, Zhengyou},
  year={2019}
}

@article{kingma2014adam,
  title={Adam: A method for stochastic optimization},
  author={Kingma, Diederik P and Ba, Jimmy},
  journal={arXiv preprint arXiv:1412.6980},
  year={2014}
}

@article{hussein2017imitation,
  title={Imitation learning: A survey of learning methods},
  author={Hussein, Ahmed and Gaber, Mohamed Medhat and Elyan, Eyad and Jayne, Chrisina},
  journal={ACM Computing Surveys (CSUR)},
  volume={50},
  number={2},
  pages={1--35},
  year={2017},
  publisher={ACM New York, NY, USA}
}

% If natbib
% \bibliography{refs}

% if IEEE
% \bibliographystyle{IEEEconf}
% \bibliography{IEEEabrv,refs}

% Or do it manually
% \begin{thebibliography}{5}

% \end{thebibliography}

%                                   % the textheight of the last page by a suitable amount.
%                                   % This command does not take effect until the next page
%                                   % so it should come on the page before the last. Make
%                                   % sure that you do not shorten the textheight too much.

% %%%%%%%%%%%%%%%%%%%%%%%%%%%%%%%%%%%%%%%%%%%%%%%%%%%%%%%%%%%%%%%%%%%%%%%%%%%%%%%%

% %%%%%%%%%%%%%%%%%%%%%%%%%%%%%%%%%%%%%%%%%%%%%%%%%%%%%%%%%%%%%%%%%%%%%%%%%%%%%%%%

% %%%%%%%%%%%%%%%%%%%%%%%%%%%%%%%%%%%%%%%%%%%%%%%%%%%%%%%%%%%%%%%%%%%%%%%%%%%%%%%%

% %%%%%%%%%%%%%%%%%%%%%%%%%%%%%%%%%%%%%%%%%%%%%%%%%%%%%%%%%%%%%%%%%%%%%%%%%%%%%%%%

\end{document}